# Design and manufacture of edible microfluidic logic gates

Shuhang Zhang[1*], Bokeon Kwak[1*], Dario Floreano[1], *Senior Member, IEEE*

*Abstract*— Edible robotics is an emerging research field with potential use in environmental, food, and medical scenarios. In this context, the design of edible control circuits could increase the behavioral complexity of edible robots and reduce their dependence on inedible components. Here we describe a method to design and manufacture edible control circuits based on microfluidic logic gates. We focus on the choice of materials and fabrication procedure to produce edible logic gates based on recently available soft microfluidic logic. We validate the proposed design with the production of a functional NOT gate and suggest further research avenues for scaling up the method to more complex circuits.

## I. INTRODUCTION

In robotics, there is an emerging trend toward renewable and biodegradable materials [1]. Among the various types of green materials, edible materials offer robots the potential of being consumed by means of deployable robots to nourish endangered animals in nature and interactive robotic cuisines for recreational dining experience. Recently, a number of edible robotic devices have been described, such as electronics [2], actuators [3], [4], and self-powered devices [5]. In related research, focus has also been on novel culinary experiences, such as shape-changing food in response to different cooking conditions [6], [7] and fluidic logic gates for flavor manipulation of desserts [8].

Control circuits are necessary to provide robots with selective and complex stimulus-response properties. Although various edible or biodegradable electric technologies have been reported, such as conductive materials [9], [10] and transistors [11], [12], electric-powered edible control units are expected to have limited driving ability due to the electric current limitation (at µA level) in edible transistors, and edible power sources [13], [14] don't yet offer the energy density necessary to drive edible electrical controllers and actuators [3] in an integrated system.

Fluid-based control circuits represent an alternative and promising way to enable complex behavior of soft robots [15]. Such fluid circuits could for example be powered using edible chemical reactions [16]. However, the use of conventional elastomeric materials (e.g., polydimethylsiloxane; PDMS) makes the circuit inedible. One edible fluid logic circuit on a centimeter scale has been reported. Herein, the logic gates are integrated within candies with the aim of modifying flavor by

This work was supported by the European Union's Horizon 2020 research and innovation program under Grant agreement 964596 ROBOFOOD. (Corresponding author: Dario Floreano)

[1]All authors are with the Laboratory of Intelligent Systems, School of Engineering, École Polytechnique Fédérale de Lausanne, CH1015 Lausanne, Switzerland (e-mail: shuhang.zhang@epfl.ch; bokeon.kwak@epfl.ch; dario.floreano@epfl.ch).

*SZ and BK contributed equally to this work.

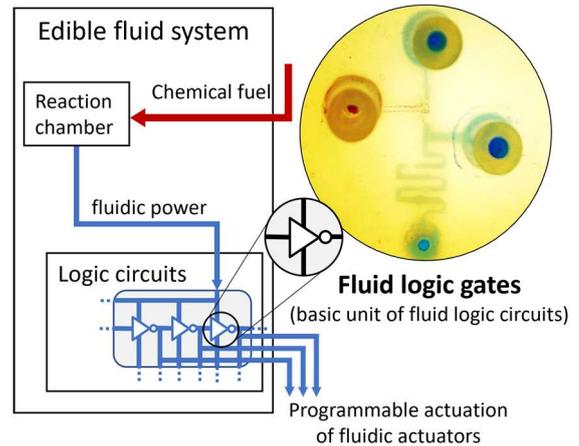

Fig. 1. Schematic representation of future edible fluidic control systems powered by chemical reaction. Red and blue lines represent chemical fuel used for gas generation and the working media fluid inside the control system respectively. This work will contribute to the fabrication procedure of edible fluid logic gates, which can be extended to building future fluidic controllers.

tailoring ingredient mixing, mediated by the logic gates [8]. However, the logic gates in this case are operated manually and not suitable for autonomous control purposes. Although the edible valve in [5] allows autonomous motion of edible fluidic actuators, digital logic behavior was not realized due to the limitation of its working principle. For future development of edible robotic foods mentioned before, more automatic and scalable edible fluidic logic circuits can be beneficial.

Here, we present a design method for the construction of microfluidic logic devices using edible materials. Edible logic circuits are the key mediator between an edible energy source (chemical fuel) and edible actuators (Fig. 1), required in the development of edible robots. Microfluidic control has been well studied and used in soft robotics for years, but the materials used to date are inedible [15], [17], [18]. We describe a material selection and fabrication procedure for edible microfluidic logic gates and validate it through the manufacturing of a NOT gate. A NOT gate was chosen as the initial proof of concept logic gate in this study because it has a convenient design, which can be implemented into more complicated logic gates and fluidic circuits [19].

## II. FLUID LOGIC DESIGN

Fluid-based logic comes in different sizes. At the macro scale (above millimeter size), several fluid-based control circuits have been built based on different pressure control valve designs [20], [21] and have been integrated into soft robots [22]. At the micro scale (below millimeter down to micrometer size) microfluidic valve designs include examples, such as the quake valve [23] and diaphragm valve [24]. Microfluidic-based control logic generally displays a flat structure, usually composed of several layers, useful for many applications. In recent work, a microfluidic controller library was built for general control purposes in soft robotics, including functions like oscillation, data storage and multiplexing [17].

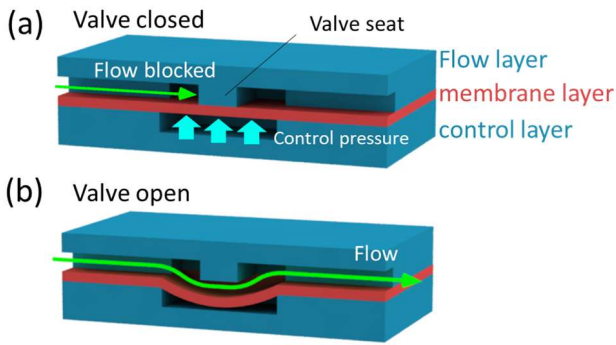

Fig. 2. Illustration of the structure of a diaphragm microfluidic valve. (a). The valve is closed given sufficient control pressure in the control layer. (b). The valve opens when control pressure is decreased, allowing the fluid to pass.

Most fluid-based logic circuits rely on external pressure sources. Aqueous solution or gas are the most common working media inside the fluid systems. Macro-scale fluid controllers require larger power input and thus have higher pressure output capability, while microfluidic controllers can work at a relatively low power consumption level. For microfluidic controllers, both vacuum [25] and positive pressure [17] power sources can be used. We chose to use positive pressure in this work to match the positive pressure generated by edible chemical reactions [16]. This will allow for easier integration of such a power source into more complex edible logic circuits in the future.

In this paper, we focus on the design of edible fluidic logic at the micro scale because microfluidic controllers have been shown to be suitable for controlling soft robots powered by chemical reactions [15]. Micro-scale controllers also require less energy than their macro counterparts and offer the potential to be integrated into smaller robotic devices that could, for example, be swallowed for medical applications. The basic component of a fluidic circuit is a valve. Here we chose to make an edible valve based on the commonly used diaphragm microfluidic valve design, which is composed of three layers (Fig. 2), [24]. These valves allow more distinguishable, binary switching compared to quake valves [23], [24], which makes them more suitable for programmable logic, while their structure is relatively simple.

The diaphragm valve comprises 3 layers: a flow layer, a membrane layer and a control layer (Fig. 2). The membrane layer is made of a flat and flexible material that blocks the flow layer when sufficient pressure is applied to the control layer. The channel in the flow layer has 2 segments, separated by an extruded part, which is called a "valve seat" (Fig. 2a). The channel in the flow layer is coupled with the channel in the control layer through the deformable membrane. The 3 layers are securely sandwiched and bonded together except for the part of the membrane adjacent to the valve seat.

When pressure is applied to one segment of the flow layer channel, the membrane layer tends to deform and open the valve. When a control pressure is applied to the control layer channel (light blue arrows in Fig. 2a), the membrane layer tends to be pushed against the valve seat to close the valve. The final state of the valve is dependent on the relationship of the two pressure values. When the flow layer pressure exceeds the control pressure, the valve will be able to open, connecting two segments of the flow channel and allowing the pressure-driven flow to pass through the flow layer (green arrows in Fig. 2b). When the control pressure is sufficiently high to overcome the pressure in the flow layer, the valve will be closed, blocking the flow in flow layer.

### III. Edible Materials for Microfluidic Circuits

The edible material used for the valve should have mechanical properties comparable to existing counterparts, since the valve works by the same mechanical principle. Here we chose to emulate PDMS, a well-established material for microfluidic logics. Two basic strain-stress parameters, tensile strength (TS) and elongation at break (EB) of edible materials are mainly considered to ensure the mechanical integrity of edible fluidic circuit and ease of analysis. The material should also have limited water affinity (i.e. solubility and swelling), which allows potential usage of water based chemical solution inside the channels. The material of choice should also allow engraving or molding to create patterns for microfluidic channels.

In addition, it is desirable for the edible material to be soft to facilitate chewing, transparent for easier inspection of fluid flow, and to have the potential of providing caloric content for nutritional value if needed.

Given the above-mentioned requirements, we took a deep look into the world of food materials. Edible hard fats, such as cocoa butter and edible waxes (carnauba wax, bees wax, etc.) are highly waterproof materials [26], solid at room temperature, and can be melted and molded into desired shapes. However, they are brittle and can't form the strong membranes required for repeated actuation of the valve [27].

Gelatin is a commonly used protein material in edible robots due to its elastic property and ease of fabrication [4], [28]. It is rich in amino acids [29] and can form durable membranes. Gelatin hydrogel has already been used for microfluidic scaffold materials [30], but is permeable to water and thus cannot be used in combination with water-soluble chemical fuels, losing or absorbing moisture in the environment also causes the channels to distort. The same problem applies to other edible polymer hydrogels, such as calcium alginate [31]. It has been reported that hydrophobic coatings can be sprayed onto the surface of gelatin to address its vulnerability to water [26], but coating of complex microchannels remains a significant challenge due to the intricacy of such a process.

Some types of edible biopolymers are hydrophobic, making them suitable for applications involving aqueous chemical solutions. These biopolymers can form networks that hold oils, which are called oleogels [32] – an analogous principle to hydrophilic biopolymers, such as gelatin, holding water to form hydrogels, such as gelatin hydrogel. For example, zein protein, which is a relatively hydrophobic protein polymer derived from corn, can form oleogel [33]. It has been shown that zein oleogel can be used to construct microfluidic devices [34] with a single microchannel layer structure. However, to achieve logic function, a multi-layer structure, which comprises microfluidic valves and a flexible membrane is needed, as described in Section II. Since zein is a brittle material even in the form of oleogel, it cannot be directly used to build such a complex structure and hence no logic function was achieved in [34]. Nevertheless, another study showed that adding a plasticizer, such as tributyl citrate, can increase the flexibility of zein [35] (decrease TS and increase EB), making it a promising option for the construction of microfluidic valves.

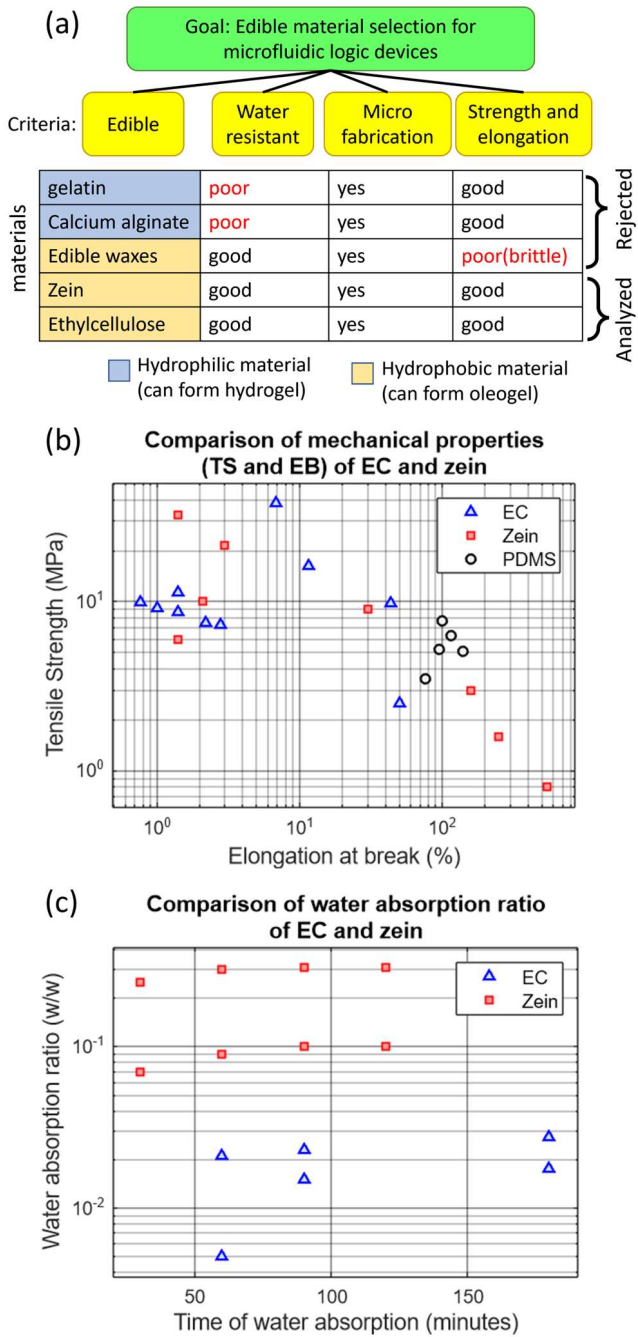

Fig. 3. Selection method and data comparison of possible materials for edible microfluidic circuits. (a). The comparison method for selecting a suitable structural material, according to the required criteria. (b), (c). Further analysis of the two main candidate materials: ethyl cellulose (EC) and zein. Data source: [35], [39]–[44]. The properties of EC and zein vary significantly by polymer source, additives and composition ratio.

Ethyl cellulose (EC) is another type of hydrophobic edible polymer able to form oleogels. It is a naturally occurring material in plants and hence environmentally friendly [36]. Owing to its excellent film-forming ability, EC has been extensively used in pharmaceuticals as a coating agent for tablets. In addition, the mechanical properties (i.e., TS and EB) of EC can be tailored by adding edible plasticizer, such as dibutyl sebacate (DBS) [37] to achieve properties, which are in line with those of silicone materials like PDMS. Although EC itself is not considered nutritious [36], it can form oleogel with edible oils, which provide fat and calories [38].

After comparing edible material candidates according to the selection criteria defined in this section (Fig. 3a), two promising materials: zein and EC were selected and further analyzed. Based on literature data [35], [39]–[44], the mechanical properties TS and EB of zein and EC membranes (containing varying amounts of plasticizer) were quantitatively compared with those of PDMS (Fig. 3b). We found that the properties of both materials can be tuned by adding plasticizer to afford similar TS and EB properties as typical PDMS. We then compared the water resistance of zein and EC films by measuring their water absorption ratio from the literature (Fig. 3c) [40], [43]. We found that the water absorption ratio of zein films (with or without plasticizer) is generally an order of magnitude higher than that of EC films, and water absorption has a significant effect on the mechanical properties of zein composites [40]. Thus, to obtain higher impermeability and structural stability in water, we chose EC as the base material for the edible microfluidic device constructed in this study (recall Fig. 2). DBS, an edible plasticizer, was added to EC to make it more stretchable. The flow (top) and control (bottom) layers of the microfluidic device were formed by adding olive oil to the DBS-containing EC, giving an oleogel, providing both nutrition and softness. The membrane layer of the device was simply manufactured from DBS-containing EC and contained no oil. Extra virgin olive oil was used due to its high nutritional and antioxidative properties, as well as heat stability, which minimizes the formation of harmful chemicals during heating at the temperatures employed in this study [45].

## IV. MANUFACTURE OF EDIBLE MICROFLUIDIC CIRCUITS

The first step towards manufacturing edible microfluidic circuits consists of creating the flow and control layers of the valve. Since EC oleogel is a thermoplastic material, replica molding is a suitable method to produce both layers (Fig. 4). To employ this method, a master mold is first needed, which is prepared by laser engraving microfluidic channel features onto an acrylic substrate [46], [47] (Fig. 4a). Compared to soft lithography [48] or micro 3d printing [49], laser engraving allows faster prototyping and was hence selected in this work. After creating the master mold, a replica silicone rubber mold (Smooth-sil[TM]960) with the microchannels embossed is derived from the master mold, and used for molding the final EC oleogel device. This replica molding method has been used for many biopolymers, including calcium alginate [31] and zein [34], since the softness of silicone molds allows for a gentle and safer demolding procedure, thus protecting the relatively weaker biomaterial structures.

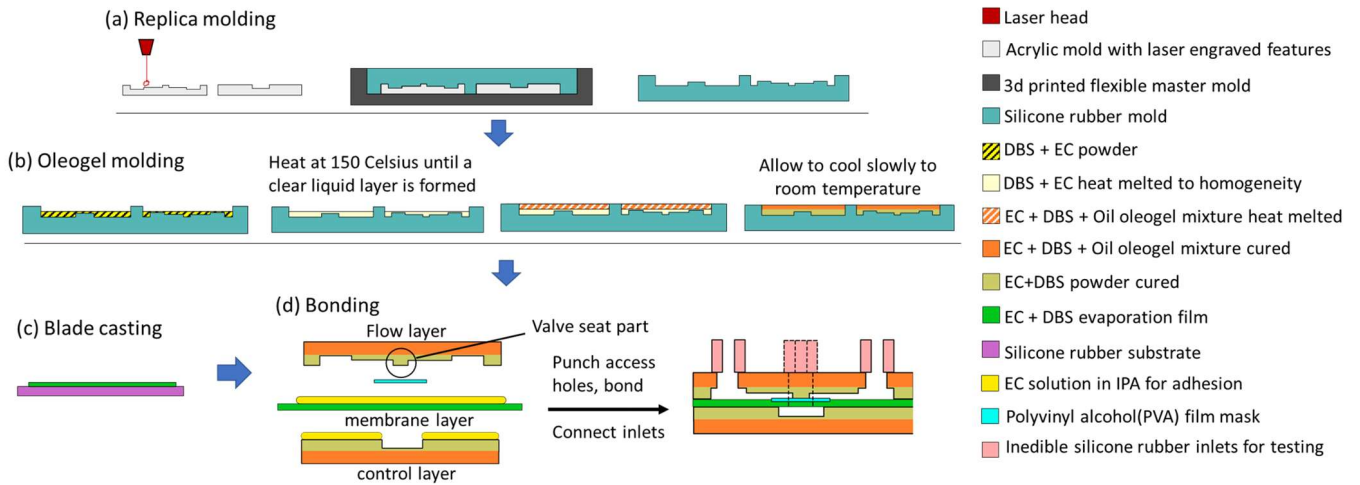

Fig. 4. Fabrication procedure of edible microfluidic logic units using selected edible material. (a). Creation of silicone rubber mold from replica molding of laser engraved features. (b). Creation of oleogel layers (flow layer and control layer) using a multilayer heat molding procedure. (c). Creation of the membrane layer using EC from solvent evaporation. (d). Bonding procedure of the layers that keeps the valve open afterwards. Dash lines represent inlets and access hole features out of the intersection plane.

Since microfluidic logic circuits are composed of multiple layer structures, which need to adhere to each other, bonding between the layers is a necessary challenge to overcome. For thermoplastic polymer materials like EC, some bonding methods are universal and interchangeable [50], [51], including adhesive bonding, solvent bonding, thermal welding, and ultrasonic welding. We found that solvent bonding using EC in isopropyl alcohol (IPA) solution works for oil-free EC layers but does not work for EC oleogel, because oil hinders bonding between the layers. Thus, to address this problem, we developed a special procedure to mold oleogel layers that would allow for bonding afterwards (Fig. 4b). Before pouring melted oleogel into the mold, we first applied a thin layer of mixed EC and DBS powder onto the mold, which was then heated to 150 °C to form a homogeneous film layer without any oil. Then we poured the homogeneous oleogel mixture (containing 1 part EC, 2 parts DBS, 3 parts oil) into the mold on top of the oil-free film. After cooling down, the molded layers were removed from the silicone mold. The EC-DBS film (which is oil-free) now works as an adhesion-promoting layer by stopping oil contamination from the EC oleogel.

Prior to the bonding of the device, the final EC membrane middle layer needed to be fabricated. A 5wt% EC solution in IPA with 5wt% DBS was blade casted onto a silicone substrate and dried under ambient condition for at least 30 minutes (Fig. 4c) repeatedly, affording a membrane with a typical thickness of 100 microns.

The device layers could then be assembled, with access holes punched through the membrane layer and flow layer before bonding Finally, the device layers were bonded using a solution of 5wt% EC in IPA (Fig. 4d). Silicone inlets were bonded onto the flow layer using inedible instant glue. These inlets were added only for testing and will not be present in future functional edible fluidic circuits. Instead, connection parts based on the same EC material can be developed to interface the logic circuits with other edible input/output components.

The final fabrication challenge was encountered when it was observed that the membrane layer could permanently adhere to the "valve seat" part during the bonding process (Fig. 4d). To address this problem, we devised a "valve seat" protection procedure, consisting of applying a mask material to the "valve seat", thus preventing its exposure to the adhesion solvent during bonding. Polyvinyl alcohol (PVA) was selected as a sacrificial mask material since it is edible, slightly soluble in the adhesion solvent (EC solution in IPA) and does not permanently bond to the EC membrane. Thus, before bonding of the flow layer and membrane layer, a piece of PVA film with a typical thickness of 30 microns is cut and placed onto the valve seat (Fig. 4d). When the bonding is complete, the PVA mask and the EC membrane are temporarily bonded. To release these two layers, water is carefully injected into the flow layer. The PVA mask is then washed away by flushing water through the flow layer. Once washing is complete, both air and water solution can be used as working media in the device.

## V. EDIBLE NOT GATE

In order to validate the materials and manufacturing process described above, we manufactured a NOT gate. A NOT gate is the most basic fluidic logic gate comprising only 1 valve, which inverts the value of its input. Any more complex logic function, such as NOR and NAND gates, is built by combining multiple NOT gates [19].

The edible NOT gate fluid circuit comprises 1 diaphragm valve and 1 pull-up fluidic resistor (Fig. 5a). It has 2 inlets that connect to a constant pressure supply and a switchable input pressure source, 2 outlets that connect to the atmosphere and an output pressure measurement point. The pressure supply is connected to the flow layer of the valve and is not altered during logic operation. The switchable pressure input is connected to the control layer of the valve: it can be turned on to block the valve and disconnect the pressure supply from the output, leading to low output pressure. When it's turned off, the flexible membrane is deflected by the supply pressure, opening the valve and allowing high output pressure to build up across the pull-up fluidic resistor. This operation corresponds to a NOT, or inverter logic operation.

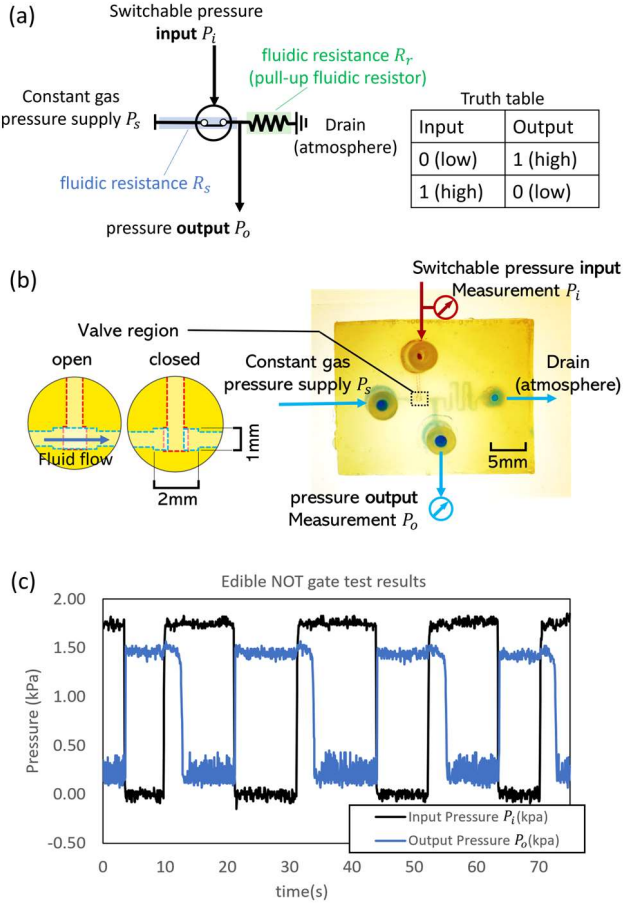

Fig. 5. Prototyping and characterization of an edible NOT gate fluidic circuit. (a). Symbolic design and truth table of a NOT gate fluidic circuit [19] as a proof of concept of the proposed fabrication method. The switch-like symbol stands for the valve. The resistor symbol stands for a pull-up fluidic resistor. (b). Photograph and zoomed-in valve illustration of a NOT gate microfluidic circuit. (c). Test result of 4 logic operation cycles from the prototype. The pressure values are calibrated relative to local atmospheric pressure during the test.

The edible NOT circuit was manufactured using the materials and methods described in the previous sections (Fig. 5b). The fabricated device has dimensions of 25mm × 20mm × 6mm, and a typical mass of 3.1 ± 0.2 g, which is dependent on fabrication error (all values were measured excluding the inedible silicone connectors). The approximate amount of fat nutrition in one device is 1.3g, providing 11.6 kcal energy. Typical height of the micro channels is 300 microns, and channel width is 0.4 mm for the fluidic resistor, 0.75mm for the remaining channels. The valve has dimensions of 1mm × 2mm. Assuming a rectangular channel cross section and air as working media, fluid resistance is calculated from these geometrical factors using the general Hagen–Poiseuille's equation for rectangular channel [52]. In this device, the fluidic resistor has a theoretical fluid resistance ($R_r$) of 8.75 × $10^8$ Pa·s/m$^3$, while the channel segment between the pressure supply inlet and the fluidic resistor (including the valve, which is seen as a fluid channel when opened) has a theoretical fluid resistance ($R_s$) of 5.89 × $10^7$ Pa·s/m$^3$ when the valve is open. With pressure supply $P_s$, the theoretical logic high value of output pressure $P_{o\_high} = P_s \cdot R_r/(R_s + R_r) = 0.92 P_s$. The input pressure $P_i$ must be greater than $P_s$ to close the valve, therefore, the highest possible control gain $P_{o\_high} / P_{i\_high}$ (ratio between $P_o$ and $P_i$'s logic high values) in theory is 0.92.

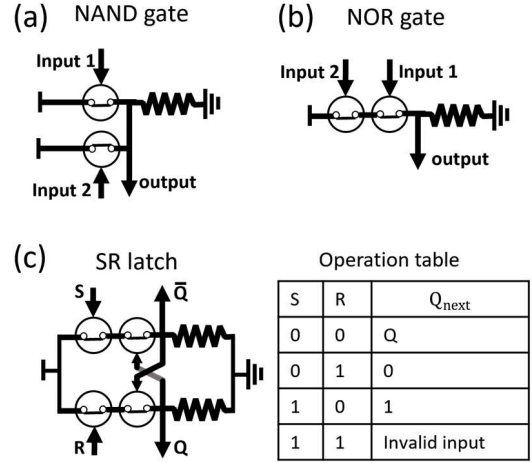

Fig. 6. Examples of future fluidic logic gate designs that could be built by combining multiple NOT gates [19]. (a), (b), NAND gate and NOR gate designs. (c) An SR latch fluidic circuit design that can hold one bit memory.

Before testing, pigmented water was injected into the device to check for leaks and help visualize the circuit. The channels were then emptied using air pressure in order to allow for the following test to take place, using air as the working medium inside the device. A wall-mounted compressed air source with a manual knob was used as the pressure source, a solenoid valve was used to achieve binary control of $P_i$. Two pressure transducers (Honeywell. Inc) were used to measure $P_i$ and $P_o$ (Fig. 5b). Note that the output measurement point is a dead end. The test system was assembled using silicone tubing.

More than 5 circuit prototypes were fabricated and tested. Here we present the results for the most successful device prototype (Fig. 5c). To start the test, the pressure source for $P_s$ is turned on and slowly increased, leading to a corresponding increase in $P_o$ measurement ($P_o$ was set at 1.46 kPa in this step). The pressure source for $P_i$ is then turned on and slowly increased to 1.77 kPa, at which point $P_o$ drops to and stays at close to 0 kPa, meaning that $P_i$ of at least 1.77 kPa is required to close the valve. Then, by turning the solenoid valve on and off, we turn $P_i$ off and on in a binary manner, leading to a corresponding, inverted binary change of $P_o$ as expected. Measurement of four typical logic state changes is shown in Fig. 5c. The logic high value of $P_i$ and $P_o$ were recorded as 1.77 kPa and 1.46 kPa w.r.t. local atmospheric pressure. Therefore, the actual control gain of the NOT gate circuit is 0.82, which is 11% lower than the theoretical value because extra $P_i$ was needed to seal the valve properly.

Noticeably, when $P_i$ was turned on, $P_o$ did not drop to its lower value immediately, but responded after a time delay of 2.7s on average. This time delay corresponded to the time needed for the valve to be fully closed by the input pressure and reach the final stable state. For other prototypes, this time delay was shorter or even close to zero, but the control gain values of other prototypes are lower than the one described above. Future research is needed to better understand the source of this time delay and thus minimize it. Fabrication consistency will also be improved in future study.

Fluidic NOT gates can be combined to build more complex fluidic logic gates such as NOR gates, NAND gates (Fig. 6a, b) and memory circuits such as the SR (Set-reset) latch (Fig. 6c) [19], such a concept has recently been demonstrated for application in interactive textile design [53]. Cascading control of logic gates is required for the construction of higher-level logic functions such as the SR latch, but a gain value smaller than 1 (0.82 for our device) makes it challenging, since the $P_o$ from one gate is not sufficient for the $P_i$ needed to control another gate driven by the same $P_s$.

There are two approaches to addressing this cascading control challenge. In [17], the problem was solved by using different pressure supplies for different logic gates. A logic gate with lower supply pressure can be easily controlled by the output of another logic gate with higher supply pressure, allowing for cascading control inside fluidic control systems. Another way to address this issue could be through the modification of the valve design. Some inedible valve designs can achieve logic gate gain values higher than 1, such as [19], [53]–[55]. For example, the gain value of a NOT gate in [19] is about 1.75, which was achieved by a modified, pre-stressed diaphragm valve design. These approaches will be tested to control cascaded edible logic gates and implement richer robotic functionalities toward edible robots and robotic foods.

## VI. CONCLUSION

We proposed a method for fabricating microfluidic logic circuits from edible materials and addressed the material selection and fabrication challenges. Compared to existing edible robotic control units, a microfluidic approach has the potential to offer better scalability and integrability for small-scale robots. We demonstrated a proof of concept NOT gate prototype, which is a crucial first step towards edible controllers and programmable motions of future edible robots. Despite the challenge of cascading control, this work forms a solid base for more complex edible micro valves and logic circuits, because our multilayer fabrication method has been designed to include all the basic procedures, such as molding and bonding, required to build more complex multilayer valves.

## ACKNOWLEDGMENT

The authors are grateful for the technical help and discussion provided by V. Annese, Y. Sun, Y. Piskarev and M. Pankhurst. This work was supported by the European Union's Horizon 2020 research and innovation program under Grant agreement 964596 ROBOFOOD.